# Multibiometrics Belief Fusion

Dakshina Ranjan Kisku
Department of Computer Science
and Engineering
Dr. B. C. Roy Engineering College
Durgapur, India
drkisku@ieee.org

Jamuna Kanta Sing
Department of Computer Science
and Engineering
Jadavpur University
Kolkata, India
jksing@ieee.org

Phalguni Gupta
Department of Computer Science
and Engineering
Indian Institute of Technology
Kanpur, India
pg@cse.iitk.ac.in

*Abstract*—This paper proposes a multimodal biometric system through Gaussian Mixture Model (GMM) for face and ear biometrics with belief fusion of the estimated scores characterized by Gabor responses and the proposed fusion is accomplished by Dempster-Shafer (DS) decision theory. Face and ear images are convolved with Gabor wavelet filters to extracts spatially enhanced Gabor facial features and Gabor ear features. Further, GMM is applied to the high-dimensional Gabor face and Gabor ear responses separately for quantitive measurements. Expectation Maximization (EM) algorithm is used to estimate density parameters in GMM. This produces two sets of feature vectors which are then fused using Dempster-Shafer theory. Experiments are conducted on multimodal database containing face and ear images of 400 individuals. It is found that use of Gabor wavelet filters along with GMM and DS theory can provide robust and efficient multimodal fusion strategy.

## I. INTRODUCTION

Recent advancements of biometrics security artifacts for identity verification and access control have increased the possibility of using identification system based on multiple biometrics identifiers [1], [2], [3]. A multimodal biometric system [1], [3] integrates multiple source of information obtained from different biometric cues. It takes advantage of the positive constraints and capabilities from individual biometric matchers by validating its pros and cons independently. There exist multimodal biometrics system with various levels of fusion, namely, sensor level, feature level, matching score level, decision level and rank level. Advantages of multimodal systems over the monomodal systems have been discussed in [2].

In this paper, a fusion approach of face [4] and ear [5] biometrics using Dempster-Shafer decision theory [7] is proposed. It is known that face biometric [4] is most widely used and is one of the challenging biometric traits, whereas ear biometric [5] is an emerging authentication technique and shows significant improvements in recognition accuracy. Fusion of face and ear biometrics has not been studied in details except the work presented in [6]. Due to incompatible characteristics and physiological patterns of face and ear images, it is difficult to fuse these biometrics based on some direct orientations. Instead, some form of transformations is required for fusion. Unlike faces, ear does not change in shape over the time due to change in expressions or age.

The proposed technique uses Gabor wavelet filters (Lee, 1996) for extracting facial features and ear features from the spatially enhanced face and ear images respectively. Each extracted feature point is characterized by spatial frequency, spatial location and orientation. These characterizations are viable or robust to the variations that occur due to facial expressions, pose changes and non-uniform illuminations. Prior to feature extraction, some preprocessing operations are done on the raw captured face and ear images. In the next step, Gaussian Mixture Model [9] is applied to the Gabor face and Gabor ear responses for further characterization to create measurement vectors of discrete random variables. In the proposed method, these two vectors of discrete variables are fused together using Dempster-Shafer statistical decision theory [7] and finally, a decision of acceptance or rejection is made. Dempster-Shafer decision theory based fusion works on changed accumulative evidences, which are obtained from face and ear biometrics. The proposed technique is validated and examined using Indian Institute of Technology Kanpur (IITK) multimodal database of face and ear images. Experimental results exhibit that the proposed fusion approach using yields better accuracy compared to existing methods.

This paper is organized as follows. Section 2 discusses the preprocessing steps involved to detect face and ear images and to perform some image enhancement algorithms for better recognition. The method of extraction of wavelet coefficients from the detected face and ear images has been discussed in Section 3. A method to estimate the score density from the Gabor responses which are obtained from the face and the ear images through Gabor wavelets has been discussed in the next section. This estimate has been obtained with the help of Gaussian Mixture Model (GMM) and Expectation Maximization (EM) algorithm [9]. Section 5 proposes a method of combining the face matching score and the ear matching score which makes use of Dempster-Shafer decision theory. The proposed method has been tested on 400 images of IITK database. Experimental results have been analyzed in Section 6. Conclusions are given in the last section.

## II. SUBJECT IMAGE LOCALIZATION AND PREPROCESSING

This section discusses the methods used to detect the facial and ear regions needed for the study and to enhance the detected images. To locate the facial region for feature extraction and recognition, three landmarks positions (as shown in Fig. 1) on both the eyes and mouth are selected and marked automatically by applying the technique proposed in [10]. Later, a rectangular region is formed around the landmarks positions for further Gabor characterization. This rectangular region is then cropped from the original face image. Original face image is constituted by facial part itself and background. For localization of ear region, Triangular Fossa [11] and Antitragus [11] are detected manually on ear image, as shown in Fig. 1. Ear localization technique proposed in [6] has been used in this paper. Using these landmarks positions, ear region is cropped from ear image. After geometric normalization, image enhancement operations are performed on face and ear images. Histogram equalization is done for photometric normalization of face and ear images having uniform intensity distribution.

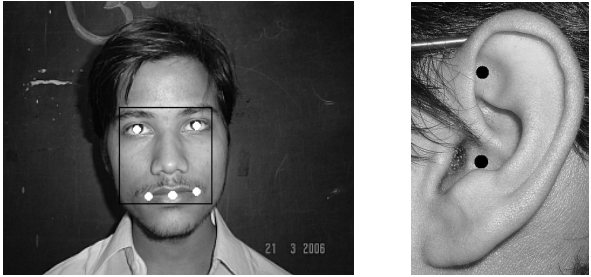

Figure 1. Geometric normalizations of face and ear images using landmark points.

## III. APPLICATION OF GABOR FILTERS TO FACE AND EAR

In the proposed approach the evidences are obtained from the Gaussian Mixture Model (GMM) estimated scores which are computed from spatially enhanced Gabor face and Gabor ear responses. Two-dimensional Gabor filter [8] refers a linear filter whose impulse response function is defined as the multiplication of harmonic function and Gaussian function. The Gaussian function is modulated by a sinusoid function. In this regard, the convolution theorem states that, the Fourier transform of a Gabor filter's impulse response is the convolution of the Fourier transform of the harmonic function and the Fourier transform of the Gaussian function. Gabor function [8] is a non-orthogonal wavelet and it can be specified by the frequency of the sinusoid and the standard deviations in both the $x$ and $y$ directions.

For the computation, 180 dpi gray scale images with the size of $200 \times 220$ pixels are used. For Gabor face and Gabor ear representations, face and ear images are convolved with the Gabor wavelets [8] for capturing substantial amount of variations among face and ear images in the spatial locations in spatially enhanced form. Gabor wavelets with five frequencies and eight orientations are used for generation of 40 spatial frequencies. Convolution generates 40 spatial frequencies in the neighbourhood regions of the current spatial pixel point. For the face and ear images of size $200 \times 220$ pixels, 1760000 spatial frequencies are generated. Infact, the huge dimension of Gabor responses could cause the performance degradation and slow down the matching process. In order to validate the multimodal fusion system Gaussian Mixture Model (GMM) further characterizes these higher dimensional feature sets of Gabor responses and density parameter estimation is performed by Expected-Maximization (EM) algorithm. For illustration, a pair of face and ear images from IITK multimodal database and their corresponding Gabor face and Gabor ear responses are shown in Fig. 2.

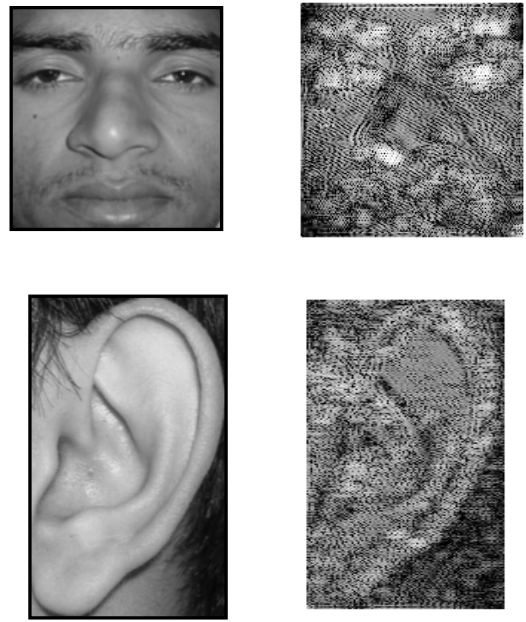

Figure 2. Face and ear images and their gabor responses.

## IV. DENSITY ESTIMATION FROM GABOR RESPONSES

Gaussian Mixture Model (GMM) [9] is used to produce convex combination of probability distribution and in the subsequent stage Expectation Maximization (EM) algorithm [9] is used to estimate the density scores. In this section, GMM is described for parameter estimation and score generation.

GMM is a statistical pattern recognition technique. The feature vectors extracted from Gabor face and Gabor ear responses can be further characterized and described by normal distributions, also called Gaussian distribution. Each quantitive measurements for face and ear are defined by two parameters: mean and standard deviation or variability among features. Suppose that the measurement vectors are the discrete random variable $x_{face}$ for face modality and variable $x_{ear}$ for ear modality. GMM is of the form a convex combination of Gaussian distributions [9]:

$$p(x_{face}) = \sum_{m=1}^{M} \pi^m p(x_{face}, \mu_{face}^{(m)}, \Sigma_{face}^{(m)}) \quad (1)$$

and

$$p(x_{ear}) = \sum_{m=1}^{M} \pi^m p(x_{ear}, \mu_{ear}^{(m)}, \Sigma_{ear}^{(m)}) \quad (2)$$

where M is the number of Gaussian mixtures and π(m) is the weight of each of the mixture. In order to estimate the density parameters of GMM, EM has been used. Each of the EM iterations consists of two steps – Estimation (E) and Maximization (M). The M-step maximizes a likelihood function that is refined in each iteration by the E-step [9].

## V. COMBINING SCORES BY DEMPSTER-SHAFER THEORY

The fusion approach uses Dempster-Shafer (DS) decision theory [7] to combine the score density estimation obtained by applying GMM to Gabor face and ear responses for improving the overall verification results. Dempster-Shafer theory is considered as a generalization of Bayesian theory in subjective probability and it is based on the theory of belief functions and plausible reasoning. Dempster-Shafer theory can be used to combine evidences obtained from different sources of system to compute the probability of an event. Generally, Dempster-Shafer decision theory is based on two different ideas such as the idea of obtaining degrees of belief for one question from subjective probabilities for a related query and Dempster's rule for fusing such degrees of belief while they depend on independent items of information or evidence [7].

Dempster-Shafer theory combines three function ingredients: the basic probability assignment function (*bpa*), the belief function (*bf*) and the plausibility function (*pf*). Let $t^{Face}$ and $t^{Ear}$ be two transformed feature sets obtained from the clustering process for the Gabor face and Gabor ear responses, respectively. Further, $m(t^{Face})$ and $m(t^{Ear})$ are the *bpa* functions for the Belief measures $Bel(t^{Face})$ and $Bel(t^{Ear})$ for the individual traits respectively. Then the belief probability assignments *(bpa)* $m(t^{Face})$ and $m(t^{Ear})$ can be combined together to obtain a Belief committed to a feature set $C \in \Theta$ according to the following combination rule or orthogonal sum rule

$$m(C) = \frac{\sum_{\Gamma^{Face} \cap \Gamma^{Ear} = C} m(\Gamma^{Face}) m(\Gamma^{Ear})}{1 - \sum_{\Gamma^{Face} \cap \Gamma^{Ear} = \varnothing} m(\Gamma^{Face}) m(\Gamma^{Ear})}, \quad C \neq \varnothing. \quad (3)$$

The denominator in Equation (3) is normalizing factor, which denotes the amounts of conflicts between the belief probability assignments *m(tFace)* and *m(tEar)*. Due to two different modalities used for feature extraction, there is an enough possibility to conflict the belief probability assignments and this conflicting state is being captured by the two *bpa* functions. The final decision of user acceptance and rejection can be established by applying threshold to m(C).

## VI. EXPERIMENTAL RESULTS

The results are obtained on multimodal database collected at IIT Kanpur. Database of face and ear consists of 400 individuals' with 2 face and 2 ear images per person. The face images are taken in controlled environment with maximum tilt of head by 20 degree from the origin. However, for evaluation purpose frontal view faces are used with uniform lighting, and minor change in facial expression. These face images are acquired in two different sessions. The ear images are captured with high-resolution camera in controlled environment with uniform illumination and invariant pose. The face and ear biometrics are statistically different from each other for an individual. One face and one ear image for each client are labeled as target and the remaining face and ear images are labeled as probe.

Table 1 illustrates that the proposed fusion approach of face and ear biometrics using DS decision theory increases recognition rates over the individual matching. The results obtained from IITK multimodal database indicates that the proposed fusion approach with feature space representation using Gabor wavelet filter and GMM outperforms the individual face and ear biometrics recognition while DS decision theory is applied as fusion rule. This fusion approach achieves 95.53% recognition rate with 4.47% EER. It has been also seen that FAR is significantly reduced to 3.4% while it is compared with the individual matching performances for face and ear biometrics. Receiver Operating Characteristic (ROC) curve is plotted in Fig. 3 for minute illustrations about the computed errors and recognition rates. The proposed fusion approach is also compared with the technique discussed in [6] and it is found to be a robust fusion technique for user recognition and authentication while the combination of Gabor wavelet filter, GMM and DS decision theory is used.

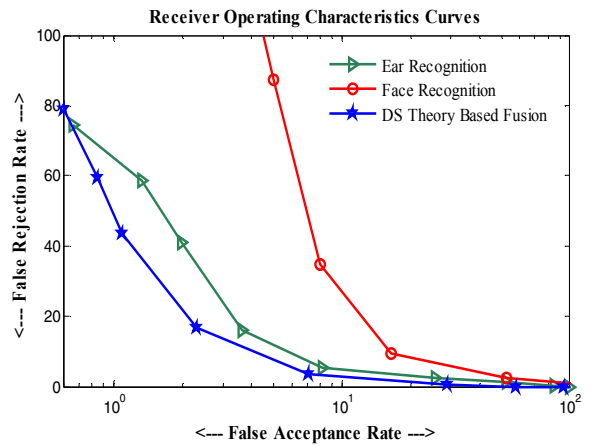

Figure 3. Receiver Operating Characteristics curves for different methods.

TABLE I. DIFFERENT ERROR RATES FOR DIFFERENT METHODS

| Methods | FRR (%) | FAR (%) | EER (%) | Recognition Rate (%) |
|---|---|---|---|---|
| Face recognition | 8.26 | 7.82 | 8.04 | 91.96 |
| Ear recognition | 7.60 | 5.70 | 6.65 | 93.35 |
| DS based fusion | 5.55 | 3.40 | 4.47 | 95.53 |

## VII. CONCLUSION

The proposed fusion strategy combines information that has been extracted through Gabor wavelet filters and Gaussian Mixture Model estimator. Gabor wavelet filters has been used for extraction of spatially enhanced face and ear features which are viable and robust to different variations. Using E-estimator and M-estimator in GMM, reduced feature sets have been extracted from high dimensional Gabor face and Gabor ear responses through parameter estimation. These reduced feature sets are fused together by Dempster-Shafer decision theory. Thorough implementation and analysis of the proposed Dempster-Shafer decision theory for fusion it has been found that the technique exhibits increase in accuracy and significant improvement over the existing methodologies.